\documentclass[twoside,11pt]{article}
\usepackage{jair}
\usepackage{rawfonts}

\jairheading{}{2023}{}{}{}
\ShortHeadings{You Don’t Need Robust Machine Learning to Manage Adversarial Attack Risks}
{Raff, Benaroch, \& Farris}
\firstpageno{1}

\usepackage{amsmath,amssymb,amsthm}
\usepackage{hyperref}

\usepackage{todonotes}

\usepackage{multirow}
\usepackage{array}
\usepackage{booktabs}

\usepackage{adjustbox}

\usepackage{todonotes}

\usepackage{natbib}
\setcitestyle{notesep={; },round,aysep={},yysep={;}} 

\begin{document}

\title{You Don't Need Robust Machine Learning to Manage Adversarial Attack Risks}

\author{\name Edward Raff \email raff\_edward@bah.com \\
        \addr 304 Sentinel Dr,\\ Annapolis Junction, MD 20701 USA
       \AND
       \name Michel Benaroch \email mbenaroc@syr.edu \\
       \addr 721 University Ave\\ Syracuse, NY 13244 USA
       \AND
       \name Andrew L. Farris \email Farris\_Drew@bah.com \\
       \addr\addr 304 Sentinel Dr,\\ Annapolis Junction, MD 20701 USA
       }

\maketitle
\begin{abstract}
The robustness of modern machine learning (ML) models has become an increasing concern within the community. The ability to subvert a model into making errant predictions using seemingly inconsequential changes to input is startling, as is our lack of success in building models robust to this concern. Existing research shows progress, but current mitigations come with a high cost and simultaneously reduce the model's accuracy. However, such trade-offs may not be necessary when other design choices could subvert the risk. In this survey we review the current literature on attacks and their real-world occurrences, or limited evidence thereof, to critically evaluate the real-world risks of adversarial machine learning (AML) for the average entity. This is done with an eye toward how one would then mitigate these attacks in practice, the risks for production deployment, and how those risks could be managed. In doing so we elucidate that many AML threats do not warrant the cost and trade-offs of robustness due to a low likelihood of attack or availability of superior non-ML mitigations. Our analysis also recommends cases where an actor should be concerned about AML to the degree where robust ML models are necessary for a complete deployment. 
\end{abstract}

\section{Introduction}

Companies are increasingly concerned with adversarial attacks to their machine learning (ML) models. In adversarial attacks, a third party wishes to subvert a company’s interests by using ML to trick their victims’ ML models to behave in a way that injures or reflects poorly on the company. Simple examples illustrate the risk. Microsoft’s Tay chatbot learned on-the-fly and was poisoned by Twitter users to produce racist tweets and producing significant backlash \citep{davis_ai_2016,wolf_why_2017}. Google’s image classifier labeled two African Americans as “gorillas” which similarly caused public outcry~\citep{barr_google_2015}. Notably, both attacks were perpetrated by regular humans attempting to confuse or find flaws in the algorithms. While these attacks are non-adversarial from a ML perspective, the significant risk of adversarial attacks stems from the thought: how dangerous could this be if vulnerability discovery is automated using ML? The risk of adversarial ML attacks is relevant to companies deploying ML models of all kinds and particularly ML security applications such as fraud, malware, and intrusion detection. Among the outcomes are fraud detection models that could be subverted, self-driving car companies that may be at risk of liability, loan applications that may create excess loss or be forced to behave in apparently discriminatory ways. Techniques for addressing or managing the risk of adversarial ML attacks are an active problem of research. 

The canonical wisdom to manage the risk of adversarial attacks is to develop so-called robust ML models  \citep{Ilyas2019}. A ML model is robust if a third party cannot reliably force the model to behave in a desired way. Robust ML models offer the benefit of being resistant (but not immune) to adversarial attacks. However, making ML models robust is non-trivial and involves a significant up-front training cost \citep{Madry2018} and ongoing cost in the form of a higher error rate \citep{Ilyas2019}. Moreover, most companies for which the risk of adversarial attacks is real are not equipped to address that risk or to develop a strategy for managing it. In a sample from 28 companies in 11 industries, only 6 companies were ready to dedicate staff to building robust ML models \citep{siva_kumar_adversarial_2020}. Simultaneously, most practitioners are completely unfamiliar with issues related to adversarial ML \citep{bieringer_industrial_2022,boenisch_i_2021}. 

We argue that immediately tackling the issue of robustness is likely counterproductive to most companies. Instead, we recommend recognizing a distinction between security and robustness in practice. Security goes beyond a ML model’s accuracy and involves the infrastructure around its maintenance, validation, and deployment that build confidence in a reliable process. Robustness of a ML model can leverage the same processes but is not just the process – it is the mechanisms by which adversarial ML attacks specifically are mitigated, and have a distinct cost. Our recommendation stems from the recognition that the need to deal with robustness is not uniform across companies or scenarios. More specifically, it stems from multiple factors about how adversarial ML works, the likelihood that an attack will be deployed, the rate of attacks, and the cost of lower accuracy robust models offer.

To justify our argument we use the following strategy: First we review the relevant related literature in \autoref{sec:related_work}, and show that the perspective of non-ML solutions to AML problems is apparently absent from the literature. Then we review the primary types of adversarial ML attacks in use today and the threat models that describe the knowledge an attacker needs to successfully perform an adversarial attack in \autoref{sec:what_is_aml}. In \autoref{sec:stylized_risk} we develop a stylized model of the cost trade-offs for the development of a robust ML model. This model then allows us to characterize, for each conjoint attack-threat model, the implied parameter values in our stylized model. By evaluating the model parameters for each attack-threat model pair, we demonstrate that adversarial attacks are less probable in most business contexts. To further justify our assumptions of risk and thus need of robust ML, in \autoref{sec:design_mitigations} we propose a number of general design choices that can be used to mitigate the risk of an AML attack. These recommendations are applied to the most prevalent and purported ``real-world'' AML attacks we found in \autoref{sec:case_studies} as case studies. Finally, \autoref{sec:who_cares} offers guidelines for contexts where pursuing robustness in ML models is warranted along with a standard operating procedure for managers to follow if adversarial ML is a risk to them. We present our conclusions in \autoref{sec:conclusion}.

\section{Related Work} \label{sec:related_work}

To the best of our knowledge, ours is the first work to discuss questions around quantifying risk from both an organizational and design perspective. Other important macro-scale aspects of adversarial attacks have been discussed but not yet connected to larger system and process designs as an effective defense.  \citep{mirsky_threat_2023} surveys the motivations and methods for how an attacker may decide and act against a victim. \citep{Brown2018} made important observations on how the threat model of an attack could be greater than what many academics consider by going beyond the standard $\|\cdot\|_p \leq \epsilon$ restrictions. \citep{zhou_adversarial_2022} mapped adversarial attacks and defenses against lessons learned and frameworks from cybersecurity, but their focus remained on attacks and defenses based on machine learning, and not alteration of a larger system or the managerial decision process in determining if the risk is acceptable.
There exist many general surveys and discussions of adversarial machine learning at many different levels, which broadly do not discuss the larger system design \citep{Yuan2017,hu_membership_2022, wang_threats_2022,li_arms_2021,Biggio2017}.

\citet{mohseni_taxonomy_2022} proposed a \textit{Taxonomy of ML Safety} by looking at a safety critical design of machine learning systems. However, we find that the survey provides no discussion on how design changes around the machine learning can mitigate the concern for attacks against the system. 
\citet{deldjoo_survey_2021} look at adversarial attacks against recommender systems. While issues about real-world problems are discussed (e.g., attacking a real-world system must factor in the temporal nature of recommendations changing over time), no practical real-world attacks are documented ``in the wild''. 
\citet{wang_threats_2022} surveyed poisoning attacks again an ML model, and did not mark any examples of this occurring in real life. Though they do mention ``Well-intentioned'' poisoning to enforce things like copyright detection (i.e., defense for something that will be made public), they do not make the full jump to recognizing poisoning as a method of countering future data theft (i.e., defense for something intended to remain private). Their survey also does not identify any recognition of how classic cryptographic key signing can be used to mitigate the risk of poisoning attacks.

\citet{hu_membership_2022} perform an extensive survey of model inversion attacks, and do note that Differential Privacy provides a provably secure method of mitigating these attacks. While they mention that DP often has a trade-off that may be too expensive, we note that it has had many successful uses in practice and thus provides a means for mitigating this class of attacks. The larger insight that a system can alter it's design to better leverage differential privacy is not discussed. 
From a different perspective \citet{paleyes_challenges_2022} focuses on how to deploy modern machine learning systems and the challenges such deployments face, with adversarial machine learning being but one concern. While they enumerate the basic attack types and some notes on the risk, they provide no guidance on how to defend against such issues from either an ML or whole system design perspective. 

We note as well that there has been limited discussion on ML from an institutional perspective in terms of maintenance and holistic design, but none that we are aware of that have tackled the heart of design changes to deal with AML risks. \citet{10.5555/2969442.2969519} talked about the technical debt of building real-world solutions, and others have talked about issues in misspecification causing inflated expectations, disappointment, and lack of trust~\citep{DAmour2020}. From a broader system design perspective of computer systems at large, seminal work by \citet{10.1145/230538.230561} discussed a number of ways bias can be introduced or emerge as a product of the larger picture (e.g., historical context, design choices, and system usage). These works share a broad theme of non-technical mitigations to technical problems, which is applicable to AML. Though we will include such design changes, we also discuss technical solutions from outside ML to the AML problem. 
Others have investigated industry-specific design concerns~\citep{kaymakci_holistic_2021}, refactoring/maintenance~\citep{tang_empirical_2021,gesi_code_2022,arpteg_software_2018}, and reproducibility~\citep{Forde2018,Raff2022a,Raff2019_quantify_repro}, but do generally focus on narrow problems and do not address larger systematic changes required to achieve technical-ML goals. 

\section{Machine Learning Models and Robustness} \label{sec:what_is_aml}

\subsection{Adversarial ML Attacks}

Adversarial examples are samples of input for a ML classification system that are very similar to a normal input example but cause the ML system to make a different classification. Adversarial examples exploit certain properties of ML classifiers and are explicitly and purposefully identified using specific algorithms called adversarial attacks. Though the mathematics of performing AML are not key to our survey as we focus on non-ML and design solutions, we briefly review them. In most AML literature this would be a $d$-dimensional feature vector $\boldsymbol{x}$ passed into a model $f(\cdot)$ for which there is a desired output $y$. The goal mean for the adversary $A(\cdot)$ who has the power to perturb the input by some $p$-norm threshold $\epsilon$, such that $\|A(\boldsymbol{x}) - \boldsymbol{x}\|_p \leq \epsilon$ and achieves the goal that $f(A(\boldsymbol{x})) \neq y$. 

Many possible targets of attack exist. In one simple scenario, attacks could allow spam or phishing attacks to go undetected by existing ML models, by forcing detection (or classification) models to make incorrect conclusions. This kind of attack can exacerbate existing cyber-security issues. In another plausible scenario, ML systems for screening credit-card charges could be fooled into classifying fraudulent transactions as non-fraudulent, allowing the adversary to cause direct financial harm and self-enrichment. Other attack scenarios could result in personally identifiable information (PII) data leaks or result in data-theft, such as replicating a company’s large investments in data labeling, warehousing, data cleaning, and model building.

It is useful to group adversarial attacks into three general types, ordered based on the nature of the risk to the enterprise \citep{siva_kumar_adversarial_2020}:
\begin{itemize}
\item Poisoning attacks seek to modify the data used to train a victim’s ML algorithm so that the attacker's goals are achieved whenever a model is trained on the poisoned data. Influencing models to have very low accuracy could amount to a “Denial of Service” attack. Poisoning could also insert “backdoors” that allow the adversary to control a model by including a special key in model input, or otherwise altering the ML model’s behavior. 
\item Inversion attacks seek to obtain information about the model itself or the data used to train it, be it by observing its behavior or by physical inspection of its parameters. They allow an adversary to create their own copy of a ML model (i.e., theft of capability) or to infer the data used in the model (e.g., PII violations and extracting individuals’ data from the model).
\item Evasion attacks trick a victim ML model into making an errant prediction due to what should have otherwise been a benign manipulation of the input data. The classic example of an evasion attack is how non-robust Computer Vision models can be fooled into making an incorrect, nonsensical prediction by altering a single pixel in the input image. 
\end{itemize}

The real-world risk of these attacks cannot be properly accounted for without considering the threat-model, which is a description of the assumptions of information required for the attack to operate successfully. The three general cases are:
\begin{itemize}
\item White-box attacks: the adversary knows everything about the victim’s ML model, including the algorithm used and any defensive techniques, and has its own copy of the training data.
\item Grey-box attacks: the adversary knows some information, but not all details. Their ability to interact with the system is limited in some ways. 
\item Black-box attacks: the adversary has only minimal access to the model, such as via an API that receives input cases and returns answers. Their ability to perturb any data is limited to data before it reaches the system. They do not know what kind of model or data is used.
\end{itemize}

\subsection{Robust ML Models}

A touted solution to adversarial ML attacks is to build robust ML models that are less susceptible to attacks. “Robustness" is a term widely used in the context of ML  to indicate that a ML system cannot be fooled by adversarial examples\footnote{e.g., see \url{https://www.robust-ml.org/}}. The exact nature of how to make ML models robust is an issue of active research. For the purposes of this survey we will use a broad definition: \textit{robust models are ones that are not easily subverted by an adversary}. One might infer that a robust model should be more accurate on all kinds of naturally occurring data and situations, though this is not commonly the case in practice. 

Obtaining robustness is non-trivial and imposes two significant costs. Making a model robust can require a significant capital expenditure. While a conventional (non-robust) ML model can cost between \$40-\$100,000 to train \citep{strubell_energy_2019}, creating a robust ML model requires expertise and significantly more training that can easily be 100 times to over 1000 times as expensive computationally \citep{Madry2018}. Moreover, when re-training must be done on regularly (e.g., on a quarterly basis) this cost is further amplified.

Another cost trade-off is reduced accuracy. Today, robust models are usually less accurate than non-robust models. Non-robust models tend to learn correlative, not causal, relationships that are brittle and thus susceptible to exploitation \citep{Ilyas2019}. These correlations are often useful in terms of predictive accuracy, but by their nature are also non-truths than an adversary can exploit. By contrast, robust models currently have lower accuracy as they forgo weak signals that are correlative but useful.

Considering the cost trade-offs presented by robust models, we posit that model robustness may be warranted only in a small set of circumstances. To make our case, we present a stylized model that shows the current best course of action for most firms is to maximize the accuracy of their ML application models and focus less on producing robust variants of their models.
\section{Stylized Model of Evasive Robustness-Security Trade
Offs} \label{sec:stylized_risk}

The following stylized model formalizes the trade-offs in cost presented by robust ML models. It enables us to compare the risk associated with adversarial attacks against a non-robust model with those against a model robust to adversarial attacks. Following this convention, we model risk exposure to an attack as
RE = (probability of an attack) × (cost consequence of the attack). 

For simplicity, assume the following parameters:

\begin{quote}
$A$ -- accuracy rate of the normal model, i.e., $(1 - A)$ is the error rate of the normal model

$p$ -- fraction of all predictions that are adversarial attacks; we pessimistically assume that all adversarial attacks on a non-robust model are successful, and optimistically assume that all adversarial attacks on a robust model are unsuccessful

$z$ -- reduction in accuracy rate in the robust model; the accuracy rate ofthe robust model is $(A - z)$ 

$(1 - (A - z))=(1 - A + z)$ -- error rate of a robust model

$c_n$ \& $c_a$  -- cost of a normal predictive error, and cost of an adversarial predictive error
\end{quote}

Ignoring the cost of training a robust model, the break-even point between a normal vs. robust model is
$\text{RE}_{\text{normal}} = \text{RE}_{\text{robust}}$, which can be expanded in terms of our assumptions as \autoref{eq:stylized_model_start} and then simplifies to \autoref{eq:stylized_model_simplified}.

\begin{equation} \label{eq:stylized_model_start}
    \underbrace{c_n (1-p) (1-A)}_{\text{Normal Errors}} + \underbrace{c_a p}_{\text{Normal Victim}} = \underbrace{c_n (1-p)(1-A+z)}_{\text{Robust Errors}} + \underbrace{c_a p (1-A+Z)}_{\text{Robust Victim}}
\end{equation}

\begin{equation} \label{eq:stylized_model_simplified}
    pc_{a}z + c_{n} = p\left( c_{a}A + c_{n}z \right)
\end{equation}

If we assume for simplicity that $c_n$ =
$c_a$, the equality simplifies to \(A\ p = z\), and the break-even condition would indicate that the cost of errors on adversarial attacks, $c_a$, must exceed the frequency of attack multiplied by the base accuracy of the model. For example, if a normal model was 95\% accurate in production use, and we believe 1\% of predictions are adversarial, a robust model must be at least
95\%-(95\%×1\%)=94.05\% accurate to be attractive to build. The more frequent attacks are, the more leniency there is to the penalty \(z\). This only considers the cost of reduced accuracy in robust models, one of the two cost trade-offs of robustness.

If we factor in the second cost trade-off and consider the training cost of a robust model, the risk exposure becomes more lopsided. A non-robust model is at risk of victimization, fraud, and other issues. A robust model presents partial mitigation to that risk and brings with it the risk that a large premium will be paid for negative net impact. Specifically, if we add the cost premium for making the model robust, denoted $D_R$, the break-even point between a normal vs. robust model becomes:

\[c_{n}\left( 1 - p \right)\left( 1 - A \right) + c_{a}p = D_{r} + \left( 1 - A + z \right)\left( c_{n}\left( 1 - p \right) - c_{a}p \right)\]

If we continue to assume that $c_n$ =
$c_a$, the break-even condition indicates that \(c_{n} = \frac{D_{r}}{Ap - z}\) the cost of errors on adversarial attacks, $c_n$, must exceed the cost premium, $D_R$, amplified by the frequency of adversarial attacks (against the baseline accuracy) modulated by the loss in accuracy \emph{z}, for a robust model to be worthwhile building. The
numerator will always be \textless{} 1, so this can only increase the costs -- and notably the penalty \emph{z} can push the cost negative, indicating that the un-satisfiability of the inequality due to the added
training costs.

The above analyses lead us to recommend against building a robust model for most companies. The benefit of a robust model is greatest when the standard model is the least accurate. This implies that the robust model will not be effective because it will be further degraded. Of course, the recommendation may be different under certain model parameters. This highlights the importance of determining the normative risks of errors when considering a robust model. More importantly, it is essential to determine if the risk is asymmetric and realistic in order to fully define the risk exposure. Next, we review these concerns in greater detail by tailoring the stylized model to various conditions.

\subsection{A Stylized Model of Cost-Benefit in Building Robust Models} 
There are two key factors that one may argue against our initial recommendation. We list these two below, so that we may further analyze the spectrum of scenarios and risk factors. 
\begin{itemize}
\item
  \emph{Cost of Adversary Attack}. It is not realistic to assume equal   costs for normal and adversary predictive errors  ($c_n$ = $c_a$). Consider an  errant approval for a loan by an ML system. In the normal context,  people may voluntarily return the money, or the legal system provides  a means to compel the return of capital (at some expense). In the
  adversarial case, an attacker may have arranged a transfer to an  uncooperative jurisdiction or arranged for money laundering via the  dark web \citep{van_wegberg_bitcoin_2018}. In such cases, the cost for errors  may differ by orders of magnitude ($c_a >> c_n$),  and thus make the development of robust predictive models viable.
\item
  \emph{Rate of Adversarial Attacks}. The rate of likely attacks and the  rate of their success may vary by the type of attack and threat model. A model trained on only publicly available data that  lacks any PII information is unlikely to be the target of theft, as  there is no competitive advantage or unique value in the data used to  construct the model. Similarly, a model that is used for internal  purposes that do not interact with any customer is less likely to be  targeted for evasion compared to a fraud model that interacts with  real (and potentially adversarial) customers.
\end{itemize}

In light of this analysis, \autoref{tbl:stylized} presents a 3×3 grid that intersects types of adversarial attacks (poisoning, inversion, and evasion) with threat models (white-, grey- and black-box) and derives, for each attack-threat model pair, implied parameter values for our stylized model. This captures the conditions for return on investment in developing robust of ML models. As seen in \autoref{tbl:stylized}, the nine possible attack/threat model combinations are categorized into four distinct groups that inform the risk analysis process. These groups are based on the viability of the attack-threat combination and tools that exist today to mitigate that threat. These categories are:

\begin{itemize}
\item
  \textbf{Realistic}: the attack could be carried out with a reasonable expectation of success. The threat-model supports the attack (i.e., the attack could happen in real life), and it can be achieved with measurable impact. A robust model would be an important defensive posture in these cases.
\item
  \textbf{Unrealistic}: The attack is not practical in most cases and unlikely to occur, absent negligence. The cost of developing a robust model is not justified.
\item
  \textbf{Solvable}: The attack could be carried out in practice, but  there are readily available techniques that can very effectively  mitigate the risk without the need to deploy a robust model.
\item
  \textbf{Impractical}: the attack could be carried out, but the  information required to perform the attack is so significant that it would present an unreasonable cost for the attacker. A robust model is unwarranted in this case because the probability of an attack is low.
\end{itemize}

\begin{table}[!h]
\centering
\caption{Table evaluating the relative risk of a stylized model of adversarial attacks. The values in each table entry correspond to the stylized model in \autoref{sec:stylized_risk}, and are inferred by the scenario and our judgment.} \label{tbl:stylized}
\begin{tabular}{@{}clll@{}}
\toprule
\multicolumn{1}{l}{} &
  \multicolumn{3}{c}{Threat Model} \\ \cmidrule(l){2-4} 
Attack Type &
  \multicolumn{1}{c}{Black-Box} &
  \multicolumn{1}{c}{Gray-Box} &
  \multicolumn{1}{c}{White-Box} \\ \midrule
Poisoning &
  \begin{tabular}[c]{@{}l@{}}p$\approx$0\\ z=Large\\ $D_R$= Large\\ $c_a$= Large\\ RE=$p \times c_a$$\approx$0\\ \textbf{Unrealistic}\end{tabular} &
  \begin{tabular}[c]{@{}l@{}}p$\approx$0\\ z= Large\\ $D_R$=Large\\ $c_a$= Large\\ RE=$p \times c_a$$\approx$0\\ \textbf{Unrealistic}\strut\end{tabular} &
  \begin{tabular}[c]{@{}l@{}}p$\approx$0\\ z= Large\\ $D_R$= Large\\ $c_a$= Large\\ RE=$p \times c_a$$\approx$0\\ \textbf{Impractical}\end{tabular} \\ \midrule
\begin{tabular}[c]{@{}c@{}}Inversion and \\ Modeling Stealing\end{tabular} &
  \begin{tabular}[c]{@{}l@{}}p=low\\ z=Low\\ $D_R$=Low\\ $c_a$=High\\ RE=$p \times c_a$ \textgreater{} 0\\ \textbf{Solvable}\end{tabular} &
  \begin{tabular}[c]{@{}l@{}}p=low\\ z=Low\\ $D_R$=Low\\ $c_a$=High\\ RE=$p \times c_a$ \textgreater{} 0\\ \textbf{Solvable}\end{tabular} &
  \begin{tabular}[c]{@{}l@{}}p$\approx$0\\ z=High\\ $D_R$=$\infty$\\ $c_a$=High\\ RE=$p \times c_a$$\approx$0\\ \textbf{Impractical}\end{tabular} \\ \midrule
Evasion &
  \begin{tabular}[c]{@{}l@{}}p=Low\\ z=Low\\ $D_R$=Large\\ $c_a$=Low-High\\ RE=$p \times c_a$ \textgreater{} 0\\ \textbf{Realistic}\end{tabular} &
  \begin{tabular}[c]{@{}l@{}}p=Medium\\ z=Medium\\ $D_R$=Large\\ $c_a$=Low-High\\ RE=$p \times c_a$ \textgreater{} 0\\ \textbf{Realistic}\end{tabular} &
  \begin{tabular}[c]{@{}l@{}}p$\approx$0\\ z=Large\\ $D_R$=Large\\ $c_a$=Low-High\\ RE=$p \times c_a$$\approx$0\\ \textbf{Impractical}\end{tabular} \\ \bottomrule
\end{tabular}
\end{table}

As shown in \autoref{tbl:stylized}, white-box attacks are generally impractical because the adversary is a powerful attacker who knows everything about the victim’s ML systems and data. Such an adversary probably has easier ways to effect negative outcomes. For example, white-box evasion attacks to alter medical imaging to change a patient’s diagnosis to/from cancer instead of benign \citep{finlayson_adversarial_2019}. While the thought is horrifying, the amount of effort required to access and alter information, undetected, in order to pull off such an attack is considerably more than simply altering a medical record to achieve the same result \citep{dr_ai_long}. Similarly, in the context of malware detection, it is easier to evade all modern anti-virus systems by employing easy, commoditized “packing” functions that obfuscate the contents of the malware, without the adversary having to rely on ML to craft undetectable malware \citep{aghakhani_when_2020}. In cases such as these, alternate methods for attack eliminate the need for the attacker to perform an adversarial ML attack. 

In practice, the difficulty of performing a real-world white-box attack is the likely reason for the lack of observed adversarial attacks in the wild. Nonetheless, white-box attacks can happen, especially when a cybersecurity incident results in a data exfiltration event \citep{nadler_detection_2019}. However, this would necessarily occur after a cybersecurity incident, making a robust model a secondary line of defense – rather than primary. In summary, the most likely conditions where a robust ML model is useful are those where a company’s IT infrastructure has already been compromised.

Compared to white-box attacks, gray and black-box attacks are more reasonable, especially when dealing with computer vision. Neural networks that are pre-trained on the publically available ImageNet dataset~\citep{he15deepresidual,ILSVRC15} are ubiquitous starting points for building computer vision systems. This makes some of the details of such a network easy to guess. Nonetheless, when researchers have evaluated the feasibility of gray-box attacks in real-world settings with imperfect knowledge, attacks are far less successful than would normally be expected. These attacks have a 33\% or lower success rate, compared to 100\% success rate in the white-box case ~\citep{Richards2021}. This significantly reduces the scope of viable attacks happening in the real world, especially when we consider that black-box attacks have even less information available. As a result, the risk of black/gray attacks depends on the type of attack. This discussion leads us to focus on gray and black-box threat models, under which evasion attacks are realistic, inversion attacks are solvable, and poison attacks are unrealistic. 

Evasion attacks are the most realistic because they require less information about the victim’s ML model. In the Evasion attack scenario, only commonly available API access is required to submit data and observe an outcome. The attacker can submit multiple queries via the API, creating an attack one step at a time. This creates a trade-off for handling the evasion attack avenue: pay the training cost premium for robustness or take on greater risk. Analytically, we see why the surprising suggestion may be to delay robustness. 

Inversion attacks, where information is leaked by the model, are also realistic but solvable. Like Evasion attacks, inversion attacks need only API access. However, tools to mitigate this risk exist for Inversion attacks. For attacks trying to obtain the original training data or extract PII information, a technique known as Differential Privacy \citep{dwork_calibrating_2006} provides a tool that is: (1) easy to add in an API, and (2) provides provable security to the results. While Differential Privacy comes at some cost of accuracy because it works by adding randomness to the process, it can be fine-tuned to balance between the extremes. Notably, Differential Privacy has been used successfully by the U.S. Census Bureau \citep{machanavajjhala_privacy_2008}, Google \citep{erlingsson_rappor_2014}, LinkedIn \citep{rogers_linkedins_2020}, and Microsoft \citep{ding_collecting_2017}. Though not a complete solution to data theft, it can help slow down the theft process \citep{cheng_differentially_2020}. In sum, under this setting, we do not see a need to add robustness to the ML model because a different tool allows us to obtain stronger guarantees as a post-processing impact with proven industry success.

Model theft through an inversion attack is problematically unrealistic. If an adversary wishes to steal a model, wouldn't it be easier to just build their own version? Stealing a model requires time and resources, combined with the fact that the stolen model will only be as good as or worse than the original. This leaves the attacker always behind the victim in model capabilities. For example, OpenAI recently created the GPT language model \citep{Radrof2019} and licensed it to Microsoft for exclusive use \$1 billion~\citep{noauthor_microsoft_2020}. Yet much of the capability was replicated by the open-source community and released for free a year later ~\citep{black_gpt-neox-20b_2022}. 

Last are poisoning attacks. In more thorough evaluations of poisoning attacks that still favor the attacker, they are shown to be often ineffective \citep{radiya-dixit_data_2022}. Similarly, there are practical options to avoiding poisoning through data oversight processes (i.e., supply-chain validation applied to your data labeling to ensure you know who is labeling and how) or rolling back to data versions before poisoning attacks became a public threat. If the attacker needs to modify significant amounts of data, their efforts are likely better spent in other ways. 

\section{Design Mitigations That Can Avoid the Need for Robust ML} \label{sec:design_mitigations}

Having elucidated a trade-off between robust and non-robust models, that is mitigated by the likelihood of an attack occurring against the average entity, we now seek to answer how such risks can be managed without relying on robust ML. This is, we argue, the most desirable outcome because it allows obtaining benefits of robustness with lesser costs. We say lesser because mitigations are not free, they could create additional friction for a user or more work for an implementer. Still, our contention is the below recommendations are better in the larger degree of certainty they provide operators to make an informed risk decision, and can confidently reduce the likelihood of attack $p$ in the stylized model from \autoref{sec:stylized_risk}. In \autoref{sec:case_studies} we will review several ``real-world'' adversarial ML attacks that largely could have been mitigated by these recommendations. Because they are meant to be general-purpose recommendations, we avoid specific scenarios unless didactic in nature.

\subsection{Poisoning Mitigation}

\subsubsection{Cryptographic Signatures of Data/Label Pairs} \label{mitigation:data_signing}

A simple strategy we have not seen discussed in the case of poisoning is to create an auditable trail of validity. That is, the threat model of poisoning attacks is often that the attacker can alter the label of your already collected data, or the content of the image. If one augments the data labeling pipeline with a cryptographic digital signature~\citep{goldwasser_digital_1988} of the tuple (data, label) a poisoning attack's likelihood becomes significantly reduced. Any alteration of the data, or the label, will result in the signature failing to validate, and thus knowledge that an attack (or data corruption) has occurred. Systems for designing and implementing key management are widely used with NIST guidance ~\citep{barker_framework_2013}. 

While it may still be plausible for an attacker to poison the source of the data, it imposes considerably stronger requirements on the adversary to be effective. Either they must:

\begin{enumerate}
    \item Create a perturbed image, which will get labeled correctly (because they alter before entry to the labeling process) and thus must require a more powerful attack to subvert a downstream model despite correct labeling. 
    \item Infect the labeling process, e.g., by being hired as a labeler, and generating sufficient bad labels to alter the results while also not getting detected as a nefarious labeler. 
    \item Somehow obtain both (1) and (2) simultaneously. 
\end{enumerate}

In all three cases, the attacker can only impact new data, which gives the defender the ability to roll back to a known good state as a further mitigation. In addition case (2) becomes increasingly more difficult if data is passed to multiple labelers to obtain better quality labels, which is a standard recommended practice ~\citep{Ratner2020,whitehill_whose_2009,NIPS2016_6523}.

\subsection{Model Inversion Mitigation}

\subsubsection{Modify Predictive Task to Enable Better Differential Privacy} \label{mitigation:dp_alter_task}

Differential privacy works best when it is naturally challenging for one datum's contents to be distinguished from others. This tends to occur with increasing frequency as the amount of data used increases, but is still susceptible to outliers in the data. An option we do not see discussed to improve the conditions of differential privacy success is to re-cast the features or predictive task used in the process. Applying normalizing transformations such as the Box-Cox transform naturally make the data better behaved to a limited distribution, and outliers or rare classes can be lumped into a single ``other'' category to increase the probability mass of a singular event (easier to make private) than many unique or extreme values (hard to keep private). While this may reduce the utility of the model, it provides a means of engineering around limitations of differential privacy. 

\subsubsection{Air-Gaped Storage of Archival vs Active Data} \label{mitigation:airgap}

Separate from the use of signed data/label pairs discussed in \autoref{mitigation:data_signing}, a further mitigation against model theft is to use an air-gapped separated between an archival store of trusted labeled data, and a working production set of data. This working set can then be \textit{defensively poisoned} or ``watermarked''~\citep{Maini2021,Song2020a,Liu2018a}, such that if theft of the data occurred via a cyber-security incident, it may be possible to identify a third party using the stolen data. 

\subsection{Evasion Mitigation}

\subsubsection{Restrict Cases Where Predictions are Made} \label{mitigation:restrict_predictions}

While a seemingly tautological argument, one can mitigate the risk of evasion attacks by predicating the prediction on a first factor. For example, requiring sign-up with a credit card, use of an RSA key to receive API access, or other barriers to use can effectively mitigate the risk of attack. The key to such cases is to make the barrier to entry a greater risk or effort to attack than the ML model itself, and it sets a new floor to the minimum amount of effort required to attack, and thus lower risk. 

Restricting predictions need not literally mean ``restrict when predictions are made via a source of friction''. Another form of restriction is to use additional non-ML and human-crafted rules, so long as they are done so with the intent to limit the ability to subvert the other rule itself. This is an intrinsically easier task to do when one is already hand-crafting a business process or rule to reason through its validity, and if it could be easily subverted, is likely not a good rule to use. 

Finally, restricting predictions can also mean restricting the useful lifetime of the predictions. Every attack intrinsically requires some amount of time to construct, and if the utility of the attack expires because the underlying model has changed in a non-trivial way, a significant barrier to attack effectiveness is created. For example, quarterly retraining of a production model is a slow means of expiring the useful life of an attack (essentially creating concept drift for the attacker), and given real-world misspecifications can dramatically reduce attack success~\citep{Richards2021}. 

\subsubsection{Audit Predictions with Gold-Label Evaluation} \label{mitigation:audit_predictions}

A practice we recommend should be done in any situation regardless of concern for adversarial attack, randomly auditing the predictions made by going through a vigorous labeling process imposes a probabilistic ceiling on the largest value of $p$, the probability of a prediction being attack, that may occur in practice.  Critically this then allows one to more empirically apply the stylized risk framework of \autoref{sec:stylized_risk}.

It is worth noting that auditing can go beyond simple input/output checks by incorporating lessons from cybersecurity \& marketing: i.e., ``know your customer''. For example, if a user of a machine-translation service appears to live in France, and is querying multiple models in multiple language pairs that do not include French, there is a greater risk they are performing some subversive behavior. By obtaining customer information and knowledge about intended and emergent use cases, errant and unusual behaviors can be flagged for follow-up to validate their authenticity. Whether the flagging results in an automated response would be a factor of risk of attack success vs user friction in using the service.

\section{Analysis of ``Real-World'' Situations} \label{sec:case_studies}

Much literature and documentation exist today on ``real-world'' cases of adversarial machine learning. If we take real-world to mean that the model/events under consideration occurred in a non-academic setting (i.e., a business, government, or organization that did not desire the events to happen) in an intended malicious fashion (i.e, an actual or risk of harm occurring with intentionality from the perpetrator), these conditions are often not satisfied. To emphasize this, we survey a number of published surveys, papers, and publicly documented examples of allegedly real-world cases of adversarial machine learning. In doing so we document a number of issues that occur that prevent this, and pair with them how to design mitigations from \autoref{sec:design_mitigations} could have alleviated risk. We categorize the potential issues that would limit the attack's risk as:

\begin{enumerate}
    \item \label{issue:easy} The work involves a threat model where the adversary must either choose to forgo an easier alternative to achieve the same goals.
    \item \label{issue:impractical} The work involves a threat model where considering a real-world motivation and goals of an attacker, the attacker's goals would not be satisfied. In such a case, there is no reason to perform the attack. 
    \item \label{issue:fixable} The defender can use existing techniques \textit{outside of machine learning} to largely mitigate the likelihood of being attacked or attack success. 
    \item \label{issue:academic} The attack has occurred in a purely academic context with curated datasets, and does not consider the scope of a full system. 
    \item \label{issue:malice} The attack was performed by academics, but in a manner simulating real world situations - or even against real production systems, but did not occur with malice. While the attack can take place, its not clear who would want to perform the attack in real-life situations, and what the actual level of risk is. 
    \item \label{issue:discoluse} The attack was performed by a third party and disclosed to the victim as a part of ``vulnerability disclosure'', and the victim took remediating actions. There are clear abilities and reasons for an adversary to perform the attack, but it might not have been confirmed in the wild. 
\end{enumerate}

Our case studies are derived primarily from MITRE Atlas case-study list of real-world adversarial attacks\footnote{\url{https://atlas.mitre.org/studies/}}. We filter from this list any case study that either: 1) does not have any other reference to the event with details. Or 2) Relied on human-only efforts to perform the attack (e.g., the Microsft Tay example). We augment this with examples of notable or highly cited works that purport to be ``practical'' or ``real-world'' examples of adversarial attacks. This leaves us with eight examples as summarized in  \autoref{tbl:cases}. We note that as shown, we do not have examples of Gray-Box inversion or Poisoning attacks, or black-box examples of poisoning attacks, despite our efforts to find examples of these situations explicitly. We speculate this is due to the difficulty of such situations as noted in \autoref{tbl:stylized}. 

\begin{table}[!h]
\caption{Summary of the threat model used for each case study. The mitigations proposed in \autoref{sec:design_mitigations} reduce the attack likelihood $p$ in all threat models. } \label{tbl:cases}
\adjustbox{max width=\textwidth}{%
\begin{tabular}{@{}lllll@{}}
\toprule
\multicolumn{1}{c}{} &
  \multicolumn{1}{c}{} &
  \multicolumn{3}{c}{Threat Model} \\ \cmidrule(l){3-5} 
\multicolumn{1}{c}{Attack Type} &
  \multicolumn{1}{c}{Mitigations} &
  \multicolumn{1}{c}{Black-Box} &
  \multicolumn{1}{c}{Gray-Box} &
  \multicolumn{1}{c}{White-Box} \\ \midrule
Poisoning &
  \begin{tabular}[c]{@{}l@{}}Cryptographic Signatures\\ of Data/Label pairs\end{tabular} &
   &
   &
  \begin{tabular}[c]{@{}l@{}}*Face Detection "hat"\\ *Facial Recognition Leak\end{tabular} \\
\begin{tabular}[c]{@{}l@{}}Inversion and\\  Model Stealing\end{tabular} &
  \begin{tabular}[c]{@{}l@{}}Modify Predictive Tasks to \\  Enable Better Differential Privacy,\\ Air-Gaped Storage of Archival\\  vs Active Data\end{tabular} &
  \begin{tabular}[c]{@{}l@{}}*Translation model Theft\\ *Search Result Copying\end{tabular} &
   &
  *Audio Speaker Verification \\
Evasion &
  \begin{tabular}[c]{@{}l@{}}Restrict Predictions Use,\\ Audit Predictions\end{tabular} &
  *Government Tax Theft &
  *Spam Filter &
  *"Good Strings" Evasion \\ \bottomrule
\end{tabular}
}
\end{table}

We now go through each case study and briefly summarize it, the issues with its realism per our six issue types, and how the scenario could have been remediated. Such remediation may not be perfect but highlights what we believe would be the most time/cost-efficient method of addressing the risk of adversarial attack. In all cases we find that robust ML methods appear to be most effective when either 1: a prior cyber-security event resulted in data theft, making white-box attacks an enhanced risk, or 2: the model requires deployment to end-users where a motivated adversary can reverse-engineer the details of the model from the deployed executable, allowing effective white-box attack. In all other situations, we find that a more thorough red-team style analysis and the mitigation strategies from \autoref{sec:design_mitigations} could be sufficient. When appropriate, we make note of any particular ``takeaway'' lesson from each case study.

\subsubsection{Malware Evasion via ``Good Strings''}

\textbf{Situation}: Anti-Virus product Cylance has a machine-learning-based detector and white-list used to avoid false positives. The white list could be reverse engineered from the product, and then tokens used by the white list inserted into malicious files. This resulted in benign predictions \citep{ashkenazy_skylight_nodate}.\\

\textbf{Issues}: \autoref{issue:discoluse}. 

\textbf{Remediation:} Robust machine learning methods are one of the only viable options in this scenario and are reported to be a part of the mitigation used. While other techniques could have helped, the fundamental issues that enable the tack are challenging to mitigate any other way. 

\textbf{Take away}: The attack works in particular because AV companies make their products available to home users, giving anyone sufficient access to perform the attack. Notably, this is also a case where better literature may have helped, as it was a rediscovery of the ``good word'' attack on spam filters that has a number of potential mitigations \citep{Lowd2005a,Jorgensen:2008:MIL:1390681.1390719,Fleshman2018a,Incer:2018:ARM:3180445.3180449}.

\subsubsection{Machine Translation Model Theft}

\textbf{Situation}: A machine-translation model can be replicated by querying the product with sentences to gain examples in another language, and then a replicate model can be trained that matches the performance of the original service \citep{wallace_imitation_2020}.\\

\textbf{Issues}:  \autoref{issue:academic}, \autoref{issue:malice}, \autoref{issue:impractical}.

\textbf{Remediation:} \autoref{mitigation:audit_predictions} None is needed as the attack considers only one pair of languages, and not the over 100 languages that such products support\footnote{\url{https://translate.google.com/intl/en-GB/about/languages/}}. Considering all pairs of languages would require running the attack over 5000 times and collecting original real sentence data for each language to then translate. At this scale of effort, there is little reason to steal the model rather than build their own translation pipeline from the ground up. 

\textbf{Take away}: The effort needed to perform an inversion attack needs to be obviously less than the effort to build a product/system to perform the same end goal in a natural manner. If there is the possibility that the attack is of comparable cost, but also carries with it a risk of legal repercussions and inability to compete long-term, then the total cost accounting for risk and opportunity cost is likely higher.

\subsubsection{Facial Recognition Dataset Leak}

\textbf{Situation}: The training data and code for a facial recognition service were obtainable by anyone because a web service was improperly configured. This created the potential for white-box attacks if previously exploited \citep{whittaker_security_2020,cameron_we_2020}.

\textbf{Issues}: \autoref{issue:discoluse}

\textbf{Remediation:} \autoref{mitigation:airgap},\autoref{mitigation:dp_alter_task}: Robust machine learning methods should become part of the solution, but risk could have been reduced by having a watermarked ``live'' dataset, with the original un-altered data air-gapped from the internet. Better design by only allowing a limited query response (i.e., ``match/no-match'') could have further reduced risk. 

\textbf{Take away}: The white-box attack threat and need for robust ML was a secondary defense caused by a lapse in basic cyber-security.

\subsubsection{Face Detection Avoiding ``Hat''}

\textbf{Situation}: A method is proposed to print out an ``adversarial patch'' (a piece of printed paper with distorted content) and place it on a hat, such that wearing the hat inhibits facial recognition systems~\citep{komkov_advhat_2021}. 

\textbf{Issues}: \autoref{issue:academic}, \autoref{issue:impractical}, \autoref{issue:easy}. 

\textbf{Remediation:} \autoref{mitigation:restrict_predictions}, \autoref{mitigation:audit_predictions}: The attack does not take into account how a system would be used, and their own results show efficacy drops significantly when applied to other recognition models. Simply using a dynamic threshold, adding cropping, or a human pre-processor to crop out the obvious sticker, would mitigate the attack. Notably, if the goal is to avoid recognition, \textit{the overt sticker itself signals to the other party that the individual is trying to hide their identity} --- and may thus draw more scrutiny than otherwise. 

\textbf{Take away}: Playing out the ``game'' of how a hypothetical larger system may react to an attack may itself mitigate the attack. By trying to avoid detection via attacking the system, the attacker may yield the inverse effect of making their presence more obvious. This observation applies to other notable works that attempt to avoid detection via ostentatious garments \citep{vedaldi_making_2020}.

\subsubsection{Spam Filter Evasion}

\textbf{Situation}: A spam filter provided by the company ProofPoint was evaded by observing meta-data the product places in emails, building a copy-cat model, and performing a transfer attack against the deployed model ~\citep{noauthor_nvd_nodate}.

\textbf{Issues}: \autoref{issue:impractical}, \autoref{issue:discoluse}. 

\textbf{Remediation:} \autoref{mitigation:restrict_predictions} None, in particular, are needed, as multiple other parts of the full detection system were not attacked \citep{noauthor_response_2020}, leaving the system as a whole still functional and low risk.

\subsubsection{China Government Tax Office Theft}

\textbf{Situation}: A real-world government facility in China used facial recognition as a means of validating invoices for payment. Attackers stole photos of other people, and used deep-fakes to mimic responsiveness and validate as the stolen party's identity to invoice the government using the stolen identity. \$77 million was fraudulently obtained \citep{olson_faces_2021}. 

\textbf{Issues}: \autoref{issue:fixable}. 

\textbf{Remediation:} \autoref{mitigation:restrict_predictions}: While not perfect, the attack could have been significantly mitigated by using more challenging biometric authentication like fingerprints. An even better option would be to require in-person registration and issuance of a cryptographic key to sign and verify identity. Multiple similar methods could be combined. 

\textbf{Take away}: A system was needlessly made more vulnerable to attack by employing facial identification via machine learning, when other alternatives are more reliable in isolation or even in combination with facial identification.

\subsubsection{Search Result Copying}

\textbf{Situation}: Google provided evidence that Microsoft's Bing was copying search results in 2011 by tracking search queries in browsers and extensions \citep{Singhal}. 

\textbf{Issues}: None. 

\textbf{Remediation:} No remediations were apparently needed. While Microsoft denies they were copied in the manner described, no legal action was taken~\citep{noauthor_google_2011}. The long-term result has been little to no discernable impact on Google's market dominance\footnote{\url{https://www.statista.com/statistics/216573/worldwide-market-share-of-search-engines/}}. 

\textbf{Take away}:Poisoning ``attacks'' are useful from the defender's perspective to gain information about whether or not the information is being stolen/used by others. However, no ML was necessary in this case from the attacker or defender, despite the subject (recommender systems via search) being an intrinsic ML problem.

\subsubsection{Audio Practical Speaker Verification}

\textbf{Situation}: A system for user verification by recognizing an individual's speech is used attacked, by creating a universal perturbation that can be played while speaking --- and fool the system into accepting a false speaker as valid~\citep{zhang_attack_2021}. 

\textbf{Issues}: \autoref{issue:academic}, \autoref{issue:fixable}, \autoref{issue:impractical}.

\textbf{Remediation:} \autoref{mitigation:restrict_predictions}, \autoref{mitigation:dp_alter_task}: The attack appears to require a room of known size and content, with a specific distance between speaker and microphone. More broadly, the attack can be defeated by white-listing to pre-registered devices (e.g., a specific phone number) that adds another layer of defense to the system. We note as well the setup already includes mitigation against leaking information about whom is authorized by requiring the use of a random phrase to be spoken, rather than a user-specified one. 

\textbf{Take away}: While impressive, the constraints apparently necessary to make the attack work in an increasingly realistic physical conditions do not inform how the larger process can be changed to mitigate threats. 

\section{When and Why Should we Focus on Robustness } \label{sec:who_cares}

Given our current analysis, it may seem that there is little reason to deal with the robustness question. This is wrong and not our message. Robustness is a challenge to implement, but it’s also a challenge for attackers to develop attacks in the real world, and the failures that must occur to make an attack more likely (data leakages or cyber security incidents) also make many adversarial attacks redundant. But, given sufficient time and resources, an adversary will be able to successfully deploy an adversarial ML attack. If a company focuses only on standard security and improved ML modeling, the adversarial attack will eventually become the lowest-effort attack vector. This means that the risk trade-off will change over time, and robustness should be on a long-term roadmap for companies at a minimum. 

That said, our results do present important themes of when an adversarial attack is an especially high risk, and so robustness of ML methods should be a primary concern. Adversarial ML attacks are a special risk for governments, for whom the adversaries are literal nation-states with enormous resources to pursue many attack vectors simultaneously, as well as banks and financial institutions for which attacks are a continuous threat due to the high potential reward. In addition, companies that produce ML models as a part of a software supply chain, where their models will be used by customers down-stream\citep{noauthor_defending_2021} and have obtained significant commercial success. While such attacks are currently a fashion of classic cyber-security issues, a realistic threat is for a hacker to compromise the supply chain and poison/steal/evade the ML models in use, so that they can stealthily influence or attack the downstream users of this system. 

This also explains why other targets, like health institutions, do not yet seem to suffer from adversarial ML attacks. Hospitals' payroll, accounts receivable, and payable, are attack vectors that may use ML-based fraud detection. But attacks today are primarily focused around ransomware~\citep{mansfield-devine_ransomware_2016,spence_ransomware_2018}, as the benefit to an adversarial attack is low, and the effort high, for an attacker today. 

For institutions that satisfy being either (1) high-value reward for attackers if successful, (2) exist as a provider in a ML supply chain, or (3) are a government entity or provider to a government entity, we recommend the following three-part strategy: 
\begin{itemize}
\item
  \textbf{Machine Learning Risk Assessments}: A critical component to   managing adversarial attacks is to develop a better understanding of the models currently used. Using the expertise of its employees or an  external contractor/specialist, companies should review applications  that integrate ML and catalog plausible attacks and the circumstances  under which these attacks may occur. A risk assessment of these models  will illuminate which attack types and threat models the client is   susceptible to. Once cataloged, a course of action can be identified  to mitigate or prevent these risks.
\item
  \textbf{Robust Machine Learning Mitigations:} Once the highest risk  models have been identified, more advanced machine learning  development can occur to improve the model's robustness. General
  purpose techniques for robustness have been improving each year, and  are currently reasonably effective to apply in cases like computer  vision~\citep{carlini_certified_2022,nie_diffusion_2022}. However, by  incorporating knowledge about the specific problem being solved, it is  possible to build significantly more robust defenses at a lower total  cost. This has been shown successfully in a defense being effective for  multiple years in computer vision ~\citep{Raff_BaRT_2019} and  for malware detection ~\citep{Fleshman2018a}.
\item
  \textbf{Extrinsic Risk Reduction:} Finally, we note that many avenues  of reducing adversarial attacks involve no machine learning at all.  Instead, changes in process or environmental factors can reduce the cost  of being attacked and the risk of attacks occurring. For example, the  aforementioned example of Microsoft's Tay chatbot was allowed to  update and redeploy the model based on live Twitter data, without any  human signoff. Instead, a process for curating the data coming in, and  reviewing new model updates before deployment, would have  significantly deterred the risk. The cost of such reviews are  ultimately minor compared to the cost of both the public relations  fallout, and the cost of developing a robust version of Tay.

Another example is that current laws allow some potential recourse, but no clear answers, on the liability and legal procedures around adversarial ML \citep{Shankar2018}. Companies can identify the changes that would simplify and support a health ecosystem of ML providers and risks so that companies can operate with confidence.
\end{itemize}

\subsection{The Intrinsic Value of Attack/Defense Research }

Beyond the risk analysis we have performed, we make special note that this should not be seen as a dismissive article against research in adversarial attacks and defenses. Indeed we argue absent any real-world attacks occurring, the questions are of a fundamentally important scientific nature. They speak to questions about intrinsic user trust in a system, that such innocuous changes can cause dramatic deviations from expectations speak to a fundamental scientific question: what do our methods learn and why. We believe and argue that this is an ever-green argument for this research direction to continue. 

\hypertarget{conclusions}{%
\section{Conclusions}\label{sec:conclusion}}

Our work elucidates that not all situations require robust machine learning to defend against adversarial attacks, and that a larger risk assessment should be performed. In real-life deployments, the cost of adding robustness may exceed its benefits. This insight is elucidated with a stylized model, from which we infer the kinds of attack scenarios that most businesses should be concerned with. To this point, we have cataloged a number of design changes that can be employed to mitigate the risks of adversarial attacks without incurring the difficulty and cost of building robust models and reviewed multiple publicly documented ``real-world'' adversarial cases to identify where these design mitigations could have been applied. In performing this analysis we have identified common themes of cases where such robustness is necessary. Our hope is that this survey will serve as a realization that the question of secure ML is broader than just ML itself, but how ML systems are designed in the context of a real-world problem to be solved. 

\bibliography{raffReferences}
\bibliographystyle{plainnat}
\end{document}